\documentclass[runningheads]{llncs}
\usepackage{graphicx}
\usepackage{subcaption}
\begin{document}
\title{Regression Concept Vectors for Bidirectional Explanations in Histopathology}
\author{Mara Graziani  $^{1,2}$
\and Vincent Andrearczyk $^{1}$  \and Henning M\"{u}ller$^{1,2}$}
\authorrunning{Mara Graziani et al.}
\institute{$^1$University of Applied Sciences Western Switzerland (HES-SO), Sierre, Switzerland \\
 $^2$University of Geneva (UNIGE), Geneva, Switzerland}
\toctitle{Lecture Notes in Computer Science}
\tocauthor{Authors' Instructions}
\maketitle
\begin{abstract}
%
Explanations for deep neural network predictions in terms of domain-related concepts can be valuable in medical applications, where justifications are important for confidence in the decision-making.
In this work, we propose a methodology to exploit continuous concept measures as Regression Concept Vectors (RCVs) in the activation space of a layer. The directional derivative of the decision function along the RCVs represents the network sensitivity to increasing values of a given concept measure. When applied to breast cancer grading, nuclei texture emerges as a relevant concept in the detection of tumor tissue in breast lymph node samples. We evaluate score robustness and consistency by statistical analysis.
\keywords{interpretability  \and concept vector \and histopathology.}
\end{abstract}
\section{Introduction}
\let\thefootnote\relax\footnote{\textit{Proceedings of the First International Workshop of Interpretability of Machine Intelligence in Medical Image Computing} at MICCAI 2018, Granada, Spain, September 16-20, 2018. Copyright 2018 by the author(s)}
Understanding representations learned by deep neural networks is a main challenge in medical imaging. Recent work on Testing with Concept Activation Vectors (TCAV) proposed directional derivatives to quantify the influence of user-defined concepts on the network output. As a real application example, the presence of diagnostic concepts such as microaneurysms and aneurysms was used to explain network predictions for diabetic retinopaty levels~\cite{KGV2017}. However, diagnostic concepts are often continuous measures that might be counter intuitive to describe by their presence or absence.

Intense research on network interpretability defined the distinction between global and local interpretability and proposed a taxonomy of desiderata, methods and evaluation criteria~\cite{DoK2017,Lip2016,MSM2017}. The relevance, or saliency, of input factors to the network decision was proposed in several gradient-based methods~\cite{MSM2017,SCD2017,SVZ2013,ZeF2013}. Outputs of these methods are typically local explanations that are gathered in attribution maps and overlayed to the original input image. The interpretability of these approaches, however, was shown to be limited and often inconsistent~\cite{KHA2017,RSG2016}. Research in the linearity of the latent space showed that linear classifiers can learn meaningful directions. These directions were mapped to semantic word embeddings in~\cite{MSC2013} or human-friendly visual concepts in~\cite{KGV2017}. TCAV computes the direction representative of a concept as the normal to the hyperplane which separates a set of concept images from a set of random images. The TCAV score estimates the influence of the user-defined concept on network decisions~\cite{KGV2017}. 

In this paper, we extend TCAV from a classification problem to a regression problem by computing Regression Concept Vectors (RCVs). Instead of seeking a discriminator between two concepts (or one concept and random inputs), we seek the direction of greatest increase of the measures for a single continuous concept. In particular, we compute RCVs by least squares linear regression of the concept measures for a set of inputs. We measure the relevance of a concept with bidirectional relevance scores, $Br$. The $Br$ scores assume positive values when increasing values of the concept measures positively affect classification and negative in the opposite case.

We address breast cancer histopathology as an application for functionally grounded evaluation. The classification of high-resolution patches as tumorous and non-tumorous tissue is often used as a first step by state-of-the-art breast cancer classifiers~\cite{GZD2017}. Identifying the factors relevant to classification is essential to improve the physicians' trust in automated grading. For this reason, we referred to the Nottingham Histologic Grading system (NHG)~\cite{ElE1991} to select nuclear pleomorphism, and especially variations in nuclei size, shape and texture as concept measures. 

The main contributions of this paper are (i) the expression of concept measures as RCVs; (ii) the development and evaluation of $Br$ scores; (iii) the computation of nuclei pleomorphism relevance for breast cancer.

\begin{figure}[ht]
    \centering
    \vspace{-.1cm}
    \includegraphics[trim={0.2cm 0.4cm 0.4cm  0.2cm},clip,width=1.\textwidth]{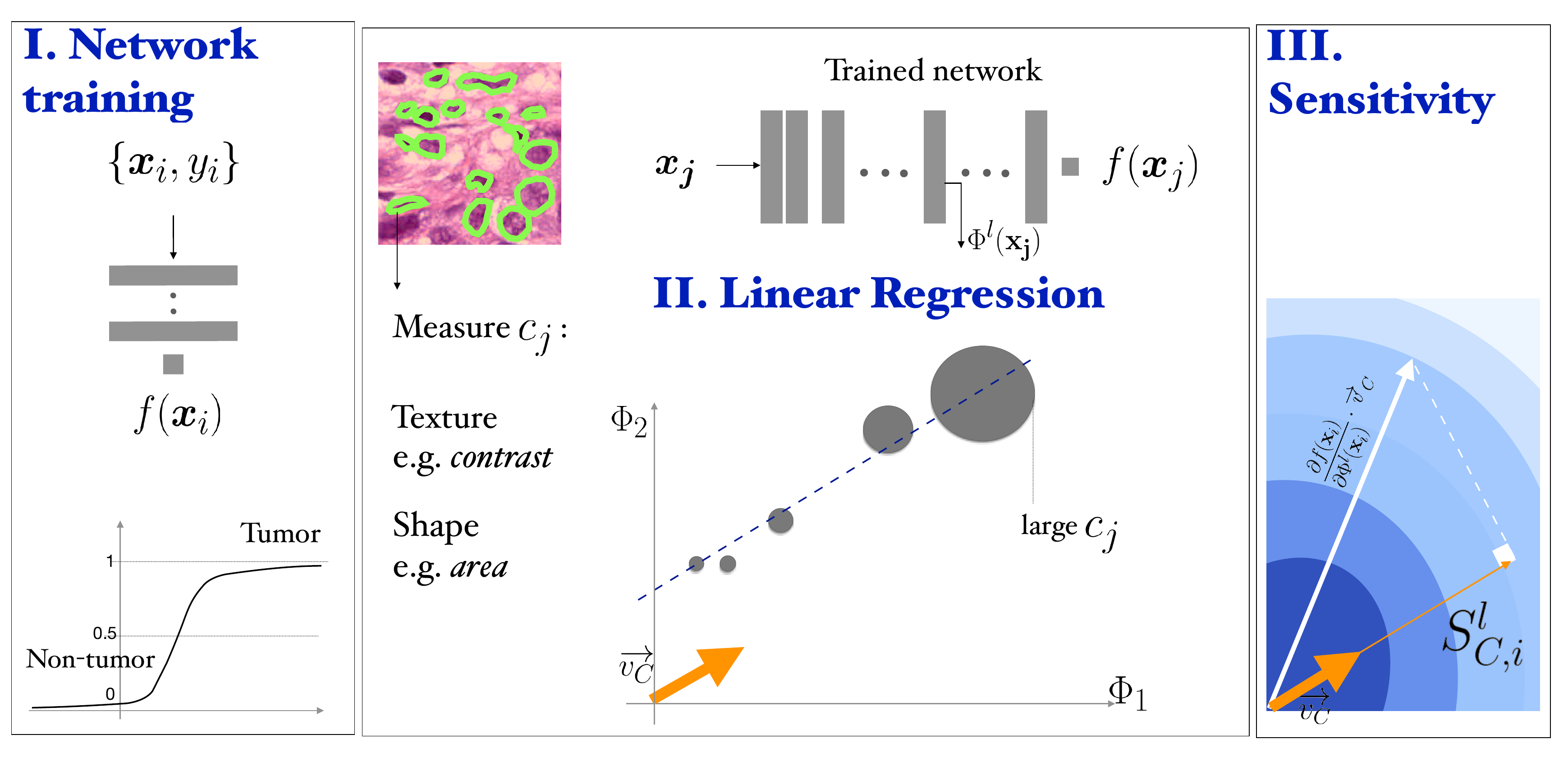}
    \caption{\textit{Method overview. I. Network training.} The last node of the CNN outputs a logistic regression function. The class \textit{Tumor} is assigned to the input patch when $f(\mathbf{x}_i)>0.5$. \textit{II. Linear Regression.} We compute average measurements of morphological and texture features from each $\mathbf{x}_j$. Linear regression $c_j=\mathbf{v}_{C}\cdot\Phi^l(\mathbf{x}_j)$ is solved on each $(\Phi^l(\mathbf{x}_j), c_j)$ at layer $l$. \textit{III. Sensitivity.} Sensitivity is computed for the $\mathbf{x}_i$ as the derivative of $f(\mathbf{x}_i)$ along $\mathbf{v}_{C}$.}
    \label{fig:overview}
\end{figure}

In the following, we clarify the notations adopted in the paper. We consider the set $\{\mathbf{x}_i, y_i\}_{i=1}^N$ of inputs and ground truth pairs and a deep convolutional neural network (CNN) for binary classification with prediction output $f(\mathbf{x}_i) \in[0,1]$. The input $\mathbf{x}_i$ is a $224\times224\times3$ image patch and $y_i \in \{0, 1\}$ is the corresponding class label (with $y=1$ for the tumor class). The disjoint set $\{\mathbf{x}_j, c_j\}_{j=1}^K$ is representative of a concept $C$, with measures $c_j \in {\rm I\!R}$ for each image sample $\mathbf{x}_j$. In the activation space, the output of layer $l$ for input $\mathbf{x}_i$ is $\Phi^l(\mathbf{x}_i)$ and the RCV for $C$ is $\mathbf{\overrightarrow{v}}_C^l$ (we will drop superscript $l$ to simplify the notation). An overview of the method is presented in Figure \ref{fig:overview}.
\section{Methods}
\label{sec:methods}
\subsection{Correlation to Network Prediction}
\label{subsec:corr}
As a prior analysis, we compute the Pearson product-moment correlation coefficient $\rho$ between $c_j$ and $f(\mathbf{x}_j)$ for $j=1,..,K$. If $c_j$ is not relevant for $f(\mathbf{x}_j)$, their correlation should be low. In this case, $\Phi^l(\mathbf{x}_j)$ should not encode information about $c_j$ and it should be unlikely to find a good linear regression. A high correlation could instead suggest a positive (if $\rho>0$) or negative ($\rho<0$) influence of the concept on the prediction.
\subsection{Regression Concept Vectors}
\label{subsec:reg_cv}
We extract and flatten the $\Phi^l(\mathbf{x}_j)$ for each $\mathbf{x}_j$. The RCV $\mathbf{\overrightarrow{v}}_C$  is the vector in the space of the activation that best fits the direction of the strongest increase of the concept measures. This direction can be computed as the least squares linear regression fit of $\{\Phi^l(\mathbf{x}_j), c_j\}_{j=1}^K$ (see Figure 1).  In the NHG, for example, larger nuclei are assigned higher grades by pathologists. If we take \textit{nuclei area} as a concept, we seek the vector in the activation space that points towards representations of larger nuclei. 
\subsection{Sensitivity to RCV}
\label{subsec:sensitivity}
For each testing pair $(\mathbf{x}_i, y_i)$ we compute the sensitivity score $S_{C,i}^l$ as the directional derivative along the direction of the RCV: 
%
%
%
%
\begin{equation}
    S_{C,i}^l = \frac{\partial f(\mathbf{x}_i)}{\partial \Phi^l (\mathbf{x}_i)} \cdot \mathbf{\overrightarrow{v}}_C
\end{equation}
$S_{C,i}^l$ represents the network sensitivity to changes in the input along the direction of increasing values of the concept measures. When moving along this direction, $f(\mathbf{x}_i)$ may either increase, decrease or remain unchanged ($S_{C,i}^l$=0). The sign of $S_{C,i}^l$ represents the direction of change, while the magnitude of $S_{C,i}^l$ represents the rate of change. TCAV computes global explanations from the $N$ sensitivities although it does not consider their magnitude. Hence, we propose $Br$ as an alternative measure. $Br$ scores were formulated by taking into account the principles of \textit{explanation continuity} and \textit{selectivity} proposed in~\cite{MSM2017}. For the former, we consider whether the sensitivity scores are similar for similar data samples. For the latter, we redistribute the final relevance to concepts with the strongest impact on the decision function. We define $Br$ scores as the ratio between the coefficient of determination of the least squares regression, $R^2$, and the coefficient of variation ${\hat{\sigma}}/{\hat{\mu}} $ of the $N$ sensitivity scores:
\begin{equation}
    Br = R^2 \times \left( \frac{\hat{\mu}}{\hat{\sigma}} \right)
\end{equation}
$R^2\le1$ indicates how closely the RCV fits the $\{\Phi^l(\mathbf{x}_i),c_i\}_{i=1}^N$. The coefficient of variation is the standard deviation of the scores over their average, and describes their relative variation around the mean. For the same value of $R^2$, the $Br$ for spread scores is lower than for scores that lay closely concentrated near their sample mean. After computing $Br$ for multiple concepts, we scale the scores to the range [-1, 1] by dividing by the maximum absolute value.
\subsection{Evaluation of the Explanations} 
\label{subsec:stat_analysis}
The explanations are evaluated on the basis of their statistical significance as proposed in \cite{KGV2017}. We compute TCAV and $Br$ scores for 30 repetitions and perform a two-tailed t-test with Bonferroni correction (with significance level $\alpha=0.01$), as suggested in~\cite{KGV2017}. If we can reject the null hypothesis of TCAV of 0.5 for random scores and Br of 0, we accept the result as statistically significant. 
\section{Experiments and Results}
\label{sec:exp}
\subsection{Datasets}
\label{subsec:datasets}
We trained the network on the challenging Camelyon16 and Camelyon17 datasets\footnote{\texttt{https://camelyon17.grand-challenge.org/} as of June 2018}. More than 40,000 patches at the highest resolution level were extracted from Whole Slide Images (WSIs) with ground truth annotation. 
To extract concepts, we used the nuclei segmentation data set in~\cite{KVS2017}, 
for which no labels of tumorous and non-tumorous regions were available. The dataset contains WSIs of several organs with more than 21,000 annotated nuclei
boundaries. From this data set, we extracted 300 training patches only from the WSIs of breast tissue.
\subsection{Network Architecture and Training}
\label{subsec:net}
A ResNet101\cite{HZR2016} pretrained on ImageNet was finetuned with binary cross-entropy loss for classification of tumor and non-tumor patches. For each input, the network outputs its probability to be tumor with a logistic regression function. We trained for 30 epochs with  Nesterov momentum stochastic gradient descent and standard hyperparameters (initial learning rate $10^{-4}$, momentum $0.9$). 
Staining normalization and online data augmentation (random flipping, brightness, saturation and hue perturbation) were used to reduce the domain shift between the different centers. 
Statistics on network performance were computed from five random splits with unseen test patients.\footnote{The pretrained models and the source code used for the experiments can be found at \texttt{https://github.com/medgift/iMIMIC-RCVs}}
\subsection{Results}
\label{subsec:res}
\begin{figure}[ht]
\centering
\begin{subfigure}{.49\textwidth}
  \centering
  \includegraphics[trim={.23cm .25cm .25cm  .25cm},clip,width=1\textwidth]{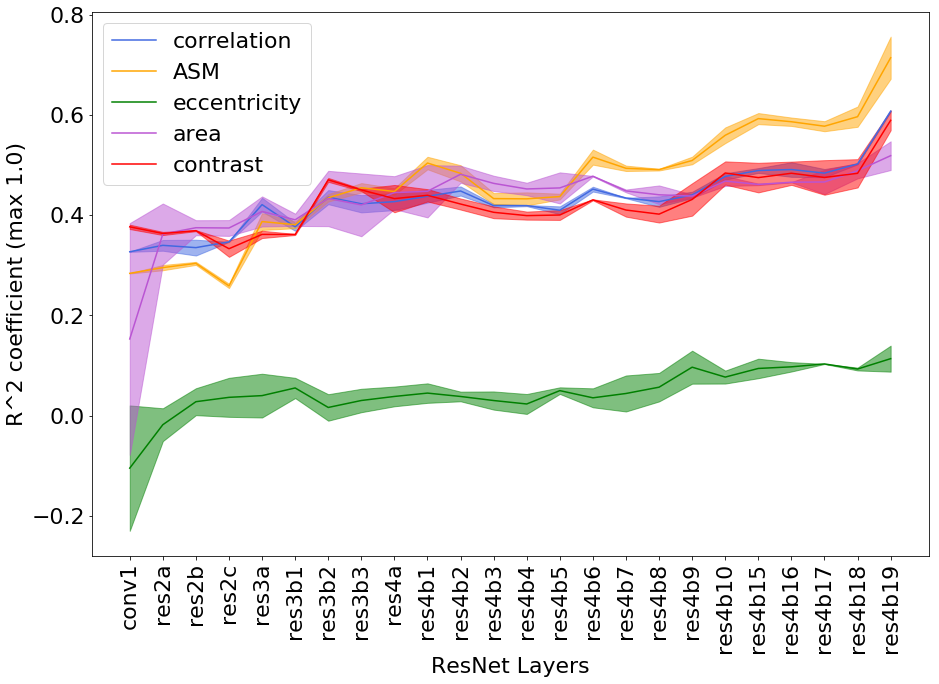}
  \caption{}
  \label{fig:stableconcepts}
\end{subfigure} \hfill
\begin{subfigure}{.49\textwidth}
  \centering
  \includegraphics[trim={.23cm 0.25cm .25cm  .25cm},clip,width=1\textwidth]{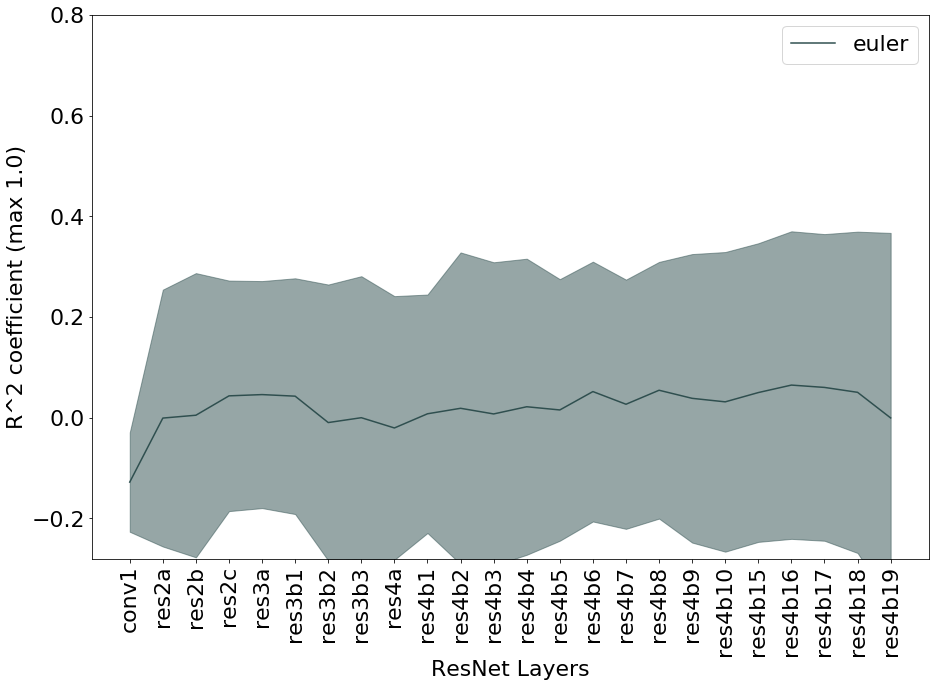}
  \caption{}
  \label{fig:unstable}
\end{subfigure}
\caption{(a)$R^2$ at different layers in the network. Results were averaged over three reruns. 95\% confidence intervals are reported. (b) The RCVs for the concept \textit{Euler} show high instability of the determination coefficient. Best on screen.}
\label{fig:detcoef}
\end{figure}

\begin{figure}[ht]
\centering
\begin{subfigure}{.45\textwidth}
  \centering
  \includegraphics[trim={.20cm 0.25cm .25cm  .25cm},clip,width=1\textwidth]{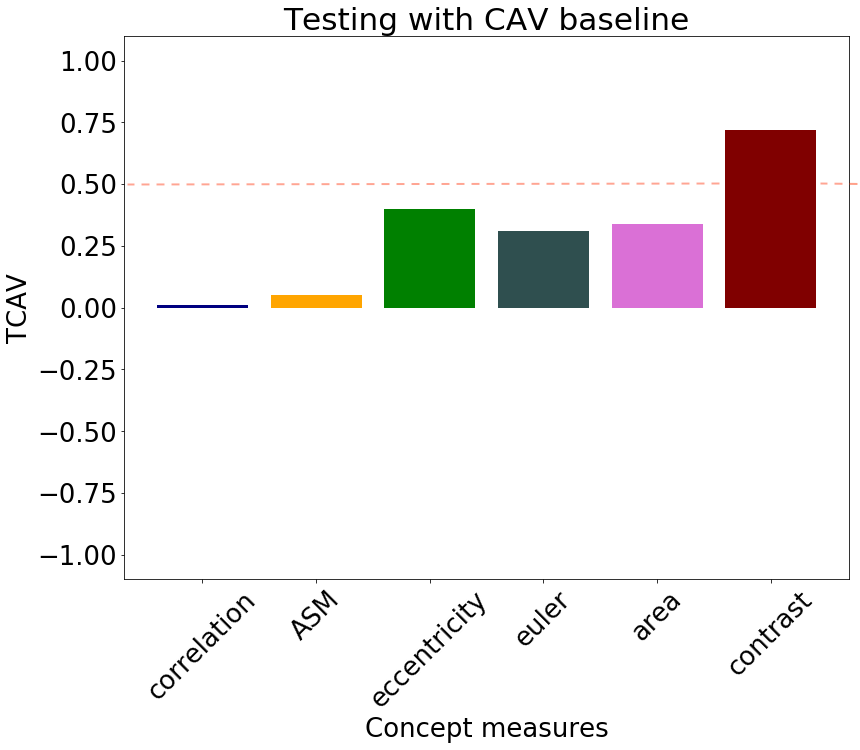}
 \caption{}
  \label{fig:a}
\end{subfigure}
\begin{subfigure}{.45\textwidth}
  \centering
  \includegraphics[trim={.23cm 0.25cm .25cm  .25cm},clip,width=1\textwidth]{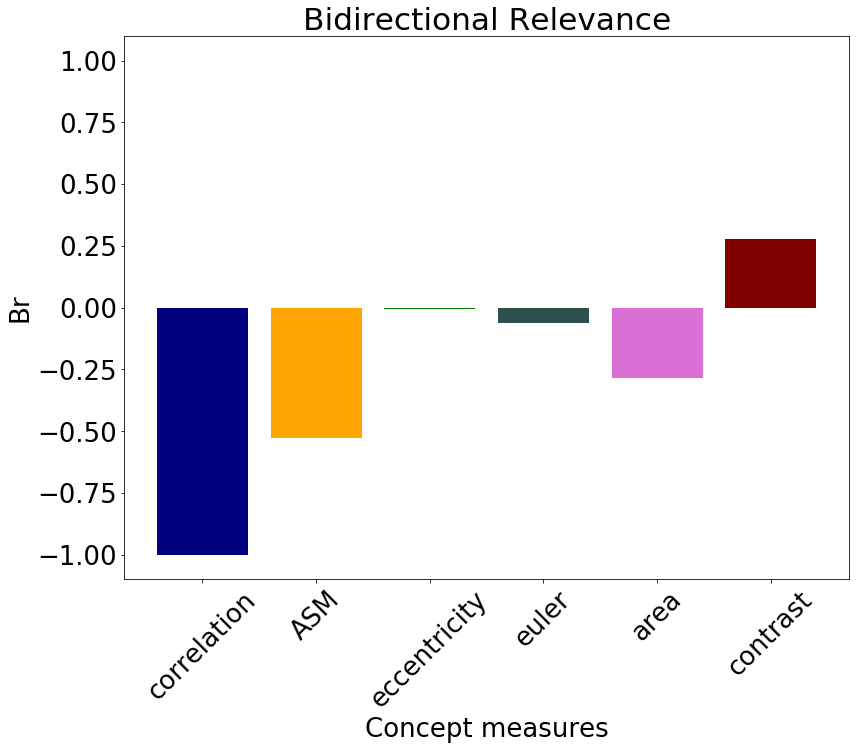}
 \caption{}
  \label{fig:b}
\end{subfigure}
\caption{Comparison of TCAV ($\in[0,1]$) and $Br$ ($\in[-1,1]$) scores. \textit{Contrast} is relevant according to both measurements. $Br$ scores show that higher \textit{correlation} drives the decision towards the non-tumor class. Scores for the unstable \textit{Euler} are approximately flattened to zero by $Br$.}
\label{fig:sensitivitsc}
\end{figure}

\paragraph{Classification Performance}
The validation accuracy of our classifier is just below the performance of the patch classifier used to get state-of-the-art results on the Camelyon17 challenge~\cite{GZD2017}, as reported in Table \ref{tab:auc}.
We report the per-patch validation accuracy for both models, although details about the training setup in~\cite{GZD2017} are unknown. Bootstrapping of the false positives was not performed and the training set size was kept limited (with 40K patches instead of 600K). The obtained accuracy is sufficient for a meaningful model interpretation analysis, which may be used to boost the network accuracy and generalization. Besides this, the analysis could itself be used as an alternative to bootstrapping for detecting mislabeled examples~\cite{KoL2017}. %
 \begin{table}[ht]
 \vspace{-.8cm}
    \centering
       \caption{Network accuracy \% for binary classification of Camelyon17 patches.}
    \label{tab:auc}
    \begin{tabular}{|c|c|c|c|c|c|c|c|c|c|c|c|c|c|}
    \hline
         model   & validation accuracy\\\hline
         Zanjani et al. & \textbf{98.7}\\
         ResNet101 & $92.43 \pm 0.657$ \\
         \hline
    \end{tabular}
\end{table}
\paragraph{Correlation Analysis}
We expressed the NHG criteria for nuclei pleomorphism as average statistics of the nuclei morphology and texture features. From the patches ($\mathbf{x}_j$) with ground truth segmentation, we computed average nuclei area, Euler coefficient and eccentricity of the ellipses that have the same second-moments as the nuclei segmented contours. We extracted three Haralick texture features inside the segmented nuclei, namely Angular Second Moment (ASM), contrast and correlation~\cite{HDS1973}. The Pearson correlation between the concept measurements and the relative network prediction is shown in Table~\ref{tab:corrtab}. The concept measures for \textit{contrast} had the largest correlation coefficient, $\rho=0.41$. 
\begin{table}[ht]
    \centering
    \vspace{-0.8cm}
        \caption{Pearson correlation between the concept measurements and the network prediction.}
    \label{tab:corrtab}
    \begin{tabular}{|c|c|c|c|c|c|c|c|c|c|c|c|c|c|}
    \hline
                    &  correlation  & ASM   & eccentricity  & Euler & area &  contrast \\\hline
         $\rho$       & {$\mathbf{-0.2985}$}   &$-0.1869$& $-0.1460$ & $0.1534$ & $0.2820$ &{$\mathbf{0.4119}$} \\
         p-value    & $\le 0.001$ & $\le 0.001$ & $\le 0.01$  & $\le 0.001$& $\le 0.001$   & $\le 0.001$ \\ \hline
    \end{tabular}
\end{table}
\paragraph{Are We Learning the Concepts?}
The performance of the linear regression was used to check if the network is learning the concepts and in which layers. The determination coefficient of the regression $R^2$ expresses the percentage of variation that is captured by the regression. We computed $R^2$ for all $x_j$ patches over multiple reruns to analyze the learning dynamics. Almost all the concepts were learned in the early layers of the network (see Figure~\ref{fig:stableconcepts}), with \textit{eccentricity} and \textit{Euler} being the only two exceptions. Figure~\ref{fig:unstable} shows that the concept \textit{Euler} is highly unstable and has almost zero mean, suggesting that the learned RCVs might be random directions.
\paragraph{Sensitivity and Relevance}
\label{sensitivityandrelevance}
Sensitivity scores were computed on $N=300$ patches ($\mathbf{x_i}$) from Camelyon17. The global relevance was tested with TCAV and $Br$, as reported in Figure \ref{fig:sensitivitsc}. \textit{Contrast} is relevant for the classification, with TCAV$=0.75$ and $Br=0.25$. Even stronger is the impact of \textit{correlation}, which shifts the classification output towards the non-tumor class. In this case sensitivies are mostly negative, with $Br=-1$ and TCAV$=0.1$. 
These scores mirror the preliminary analysis of Pearson correlation in Table~\ref{tab:corrtab}. Unstable concepts, such as \textit{Euler} and \textit{eccentricity}, lead to almost zero $Br$ scores, in accordance with the initial hypothesis that the RCVs for these concepts might just be random vectors. 
\paragraph{Statistical Evaluation}
We performed a two-tailed t-test to compare the distributions of the scores against the null hypothesis of learning a random direction for the TCAV (mean $0.5$) and $Br$ (mean $0$) scores. 
The results are presented in Table \ref{tab:statpvalues}. There was a significant difference (with $p\textnormal{-}value\leq0.01$) in the scores for all the relevant concepts, namely \textit{correlation}, \textit{ASM}, \textit{area} and \textit{contrast}. The statistical significance of \textit{correlation} improves for $Br$ scores. 
From the sensitivity and relevance analysis, we do not expect the \textit{Euler} and \textit{eccentricity} concepts to be statistically different from random directions. The analysis of both TCAV and $Br$ scores confirms this hypothesis ($p\textnormal{-}value\leq0.01$) for the \textit{eccentricity}, although the confidence to not reject the null hypothesis is higher with $Br$. The \textit{Euler} concept is not rejected by the TCAV analysis. $Br$ scores, instead, reject the hypothesis of this score being relevant. 
\begin{table}[ht]
\vspace{-0.5cm}
 \caption{Statistical significance of the scores. The p-values are reported for two-tailed t-tests evaluating the difference between the distributions of the obtained scores against a normal distribution of the scores for random concepts, i.e. mean 0.5 for TCAV and 0 for $Br$.}
   \label{tab:statpvalues}
    \centering
       \begin{tabular}{|c|c|c|c|c|c|c|c|c|c|c|c|c|c|}
    \hline
             &  correlation  & ASM   & eccentricity  & Euler & area & contrast \\\hline  
         TCAV   & $0.002$ & $ 0.001$ & $ 0.02$  & $ 0.01$& $ 0.001$  &  $ 0.001$ \\\hline
         $Br$    & $0.001$ & $ 0.001$ & $0.30$  & $1.0$& $ 0.001$  & $ 0.001$ \\\hline
    \end{tabular}
\end{table}
\section{Discussion and Future Work}
\label{sec:discussions}
RCVs showed that nuclei \textit{contrast} and \textit{correlation} were relevant to the classification of patches of breast tissue. This is in accordance with the NHG grading system, which identifies hyperchromatism as a signal of nuclear atypia. Extending the set of analyzed concepts can lead to the identification of other relevant concepts. RCVs can 
give insights about network training. The learning of the concepts across layers is linked to the size of the receptive field of the neurons and the increasing complexity of the sought patterns (see Fig. \ref{fig:detcoef} and \cite{KGV2017}). Hence, more abstract concepts, potentially useful in other applications, can be learned and analyzed in deep layers of the network. Moreover, outliers in the values of the sensitivity scores can identify challenging training inputs or highlight domain mismatches (e.g. differences across hospitals, staining techniques, etc.). 

Overall, this paper proposed a definition of RCVs and a proof of concept on breast cancer data. RCVs could be extended to many other tasks and application domains. In the computer vision domain, RCVs could also express higher-level concepts such as materials, objects and scenes. In signal processing tasks, RCVs could be used, for instance, to determine the relevance of the occurrence of a keyword in topic modeling, or of a phoneme in automatic speech recognition.
\paragraph{Acknowledgements}
This work was possible thanks to the project PROCESS, part of the European Union’s Horizon 2020 research and innovation program (grant agreement No 777533).

\bibliographystyle{plain}
%

\end{document}